\documentclass[conference]{IEEEtran}
\IEEEoverridecommandlockouts
\usepackage{cite}
\usepackage{amsmath,amssymb,amsfonts}
\usepackage{graphicx}
\usepackage{textcomp}
\usepackage{xcolor}
\usepackage{xspace}
\def\BibTeX{{\rm B\kern-.05em{\sc i\kern-.025em b}\kern-.08em
    T\kern-.1667em\lower.7ex\hbox{E}\kern-.125emX}}

\newcommand{\mm}{\textsc{ReForge}\xspace}
\usepackage{flushend}
\usepackage{CJKutf8}
\usepackage{amsmath, amsfonts}
\usepackage{array}
\usepackage{textcomp}
\usepackage{stfloats}
\usepackage{url}
\usepackage{xurl}
\usepackage{balance}
\usepackage{verbatim}
\usepackage{graphicx}
\usepackage{cancel}
\usepackage{cite}
\usepackage{graphicx}
\usepackage{multirow}
\usepackage{booktabs}
\usepackage{tabularx}
\usepackage{graphicx}
\usepackage{array}
\usepackage{tabularx}
\usepackage[caption=false,font=footnotesize]{subfig}

\usepackage{xcolor}
\usepackage{algpseudocode}

\usepackage[ruled,vlined,linesnumbered]{algorithm2e}

\usepackage{hyperref} 
\hypersetup{
    hidelinks,
    colorlinks=true, 
    linkcolor=blue, 
    citecolor=blue, 
    urlcolor=blue   
}
\newcolumntype{Y}{>{\centering\arraybackslash}X}

\usepackage{pgfplots}
\pgfplotsset{compat=1.18}
\begin{document}

\title{REFORGE: Multi-modal Attacks Reveal Vulnerable Concept Unlearning in Image Generation Models}

\author{
    Yong Zou$^{1}$, Haoran Li$^{2}$, Fanxiao Li$^{1}$, Shenyang Wei$^{1}$, Yunyun Dong$^{1}$, Li Tang$^{1}$, Wei Zhou$^{1}$, and Renyang Liu$^{*,3}$%
    \\
    $^1$Yunnan University $^2$Northeastern University $^3$National University of Singapore
    \thanks{*Corresponding author. This work is supported by the Yunnan Research Project (Grant Nos. 202503AG380006, 202401AT070474, 202501AU070059, 202403AP140021), National Natural Science Foundation of China (Grant Nos. 62562061, 62502422 and 62462067), and Yunnan Provincial Department of Education Science Research Project (Grant Nos. 2025J0006, 2024J0010 and 2025J0007). (Email: ryliu@nus.edu.sg)}
}
\maketitle

\begin{abstract}
Recent progress in image generation models (IGMs) enables high-fidelity content creation, but amplifies risks including reproducing copyrighted or generating offensive content. Image Generation Model Unlearning (IGMU) mitigates these risks by removing harmful concepts without full retraining. Despite growing attention, the robustness under adversarial inputs, particularly image-side threats in black-box settings, remains underexplored. To bridge this gap, we present \mm, a black-box red-teaming framework that evaluates IGMU robustness via adversarial image prompts. \mm initializes stroke-based images and optimizes perturbations with a cross-attention-guided masking strategy that allocates noise to concept-relevant regions, balancing attack efficacy and visual fidelity. Extensive experiments across representative unlearning tasks and defenses demonstrate that \mm significantly improves attack success rate while achieving stronger semantic alignment and higher efficiency than involved baselines. These results expose persistent vulnerabilities in current IGMU methods and highlight the need for robustness-aware unlearning against multi-modal adversarial attacks. Our code at: \url{https://github.com/Imfatnoily/REFORGE}.

\end{abstract}


\section{Introduction}

Image generation models (IGMs) have witnessed remarkable progress, revolutionizing applications in artistic creation, virtual reality, and medical imaging. Prominent models, such as DALL·E~\cite{dalle2}, Imagen~\cite{imagen}, and Stable Diffusion~\cite{ldm}, have facilitated the widespread adoption of text-to-image synthesis. However, these capabilities have introduced significant safety and compliance concerns~\cite{deepfake}, including harmful, misleading, or copyright-infringing generations that can cause tangible societal threats.

A key source of these risks is the reliance of modern IGMs on large-scale internet-scraped datasets~\cite{laion}, which inevitably contain copyrighted works, NSFW imagery, etc. Such undesirable information can be internalized during training and later re-emerge at deployment time, enabling misuse even when the service interface appears benign.

Although dataset filtering followed by retraining can mitigate these issues, it is often computationally prohibitive for large-scale diffusion models~\cite{sd2}. Consequently, prior work mainly follows two directions: (1) external filters that screen prompts or generated images~\cite{dalle2,sd_safety_checker, sd_safety_checker1}, and (2) Machine Unlearning (MU) that removes specific concepts by directly modifying model parameters~\cite{esd, uce, fmn, mace, AdvUnlearn, DoCo, ConceptPrune}. Filtering-based defenses suffer from inherent trade-offs: pre-filtering can over-lock benign prompts, whereas post-filtering increases inference latency and wastes computations on discarded generations. In contrast,  Image Generation Model Unlearning (IGMU) integrates the removal objective into the model itself, offering a more direct and potentially efficient mitigation.

\begin{figure}[t]
    \centering
    \includegraphics[width=\linewidth]{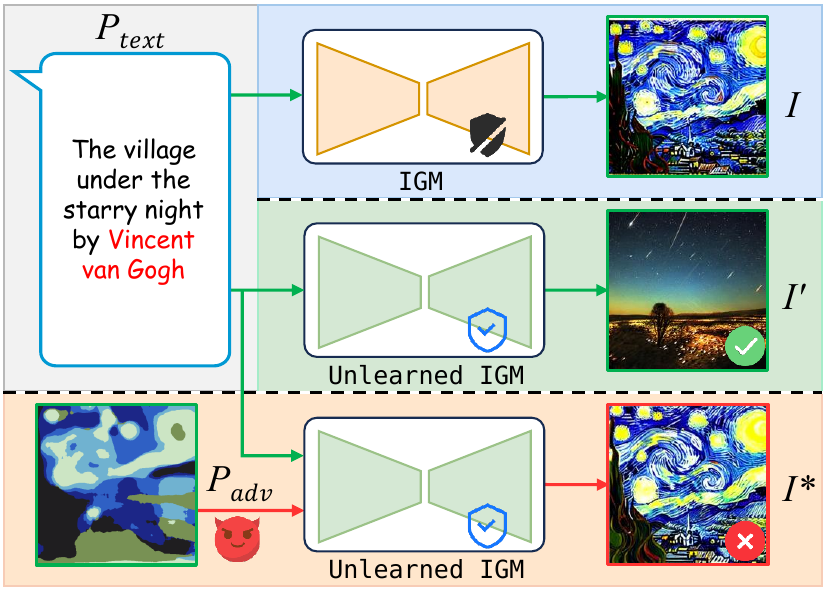}
    \caption{Given that an unlearned IGM has undergone a concept-unlearning procedure (e.g., removal of the Van Gogh style), our adversarial image prompt $P_{adv}$ combined with the prompt $P_{text}$ can still bypass the unlearning mechanism, causing the erased style to re-emerge in the generated image $I^{*}$.}
    \label{fig:cover}
\end{figure}

Researchers have developed diverse IGMU techniques, including inference-time constraints~\cite{SLD,arXiv_26/saferedir}, weight editing~\cite{esd, uce}, adversarial training~\cite{AdvUnlearn}, and structural pruning~\cite{ConceptPrune}. However, the robustness of unlearned models against adversarial inputs remains insufficiently understood. Recent studies show that erased concepts can be recovered via carefully optimized prompts. White-box attacks, such as P4D~\cite{P4D} and UnlearnDiffAtk~\cite{UnlearnDiffAtk}, exploit access to model internals to construct effective adversarial prompts. In the black-box setting, existing red-teaming methods largely focus on manipulating text prompts~\cite{SneakyPrompt, ringabell, JPA, diffzoo, JailFuzzer}, while the vulnerabilities introduced by image inputs are less explored. Although recent work~\cite{ReCall} investigates image-modality red-teaming, it relies on white-box access. To our knowledge, black-box red-teaming for IGMU image inputs remains unstudied.

To bridge this gap, we study the robustness of unlearned IGMs under realistic text-to-image generation interfaces where attackers can provide both text and image inputs. We propose \mm, a novel black-box red-teaming framework that generates adversarial image prompts to bypass IGMU mechanisms. As illustrated in Fig.~\ref{fig:cover}, \mm combines adversarial stroke-based image prompts with the original text prompt to induce the re-emergence of erased concepts while preserving overall semantic consistency. Crucially, \mm does not require access to target-model parameters or gradients, making it applicable to real-world, closed-source services.

We validate \mm through extensive experiments across three representative unlearning categories and multiple concept erasure techniques. The experimental results demonstrate that \mm achieves superior performance in terms of attack success rate, semantic similarity, and attack efficiency, compared to representative baselines. Our key contributions are as follows:
\begin{itemize}
    \item We propose \mm, a black-box red-teaming framework that targets the image modality for IGMU and reveals the fragility of current unlearning mechanisms under realistic multi-modal attacks.
    \item We introduce a masking strategy that leverages cross-attention maps to allocate perturbations, balancing attack effectiveness with visual imperceptibility.
    \item We conduct extensive evaluations across unlearning tasks and methods, showing that \mm consistently outperforms prior baselines in effectiveness, semantic preservation, and efficiency.
\end{itemize}

\section{Related Work}

\subsection{Image Generation Model Unlearning}
As IGMs improve in fidelity, they also amplify safety and compliance risks by enabling the synthesis of undesirable content. Image generation model unlearning (IGMU) aims to remove specific concepts from a pretrained generator while preserving general generation quality. Existing IGMU methods span inference-time suppression, parameter editing, adversarial optimization, and structural pruning. Specifically, SLD~\cite{SLD} imposes suppression constraints at inference time, whereas ESD~\cite{esd} performs selective fine-tuning over model layers. UCE~\cite{uce}, MACE~\cite{mace}, and RECE~\cite{RECE} use closed-form updates for efficient weight modification: UCE targets cross-attention parameters, MACE integrates LoRA modules for multi-concept erasure, and RECE iteratively eliminates derived embeddings with regularization to preserve generation quality. FMN~\cite{fmn} achieves unlearning through attention redirection. AdvUnlearn~\cite{AdvUnlearn} leverages adversarial examples to enhance forgetting robustness, and DoCo~\cite{DoCo} adopts a GAN-like framework with adversarial optimization. ConceptPrune~\cite{ConceptPrune} removes concepts by pruning critical neurons in the FFN layers of denoiser.

\subsection{Red Teaming for Image Generation Model Unlearning}
Despite progress in IGMU, recent studies have shown that erased concepts can be recovered under adversarial inputs. In white-box settings, P4D~\cite{P4D} uses an auxiliary diffusion model to optimize adversarial prompts, and UnlearnDiffAtk~\cite{UnlearnDiffAtk} improves efficiency by an additional reference image. Beyond gradient-based methods, SneakyPrompt~\cite{SneakyPrompt} adopts reinforcement learning for prompt optimization, Ring-A-Bell~\cite{ringabell} applies genetic algorithms to align prompts with concept vectors, and JPA~\cite{JPA} relaxes discrete tokens into continuous variables for efficient optimization. For black-box red-teaming, DiffZOO~\cite{diffzoo} performs zeroth-order optimization, and JailFuzzer~\cite{JailFuzzer} employs large language models as fuzzing agents.

While these efforts have substantially advanced red-teaming for unlearning, most existing frameworks operate primarily in the text modality and do not explicitly account for the image-input channel supported by many IGMs. Although RECALL~\cite{ReCall} extends red-teaming to the image modality, it relies on white-box assumptions, leaving black-box evaluation via image inputs largely unexplored. To fill this gap, we propose \mm, a black-box robustness assessment framework for multi-modal scenarios, demonstrating that erased concepts can be recovered by combining unmodified textual prompts and adversarial stroke-based image prompts. \mm does not require access to the target model’s parameters or gradients, making it applicable to real world scenarios.

\definecolor{myyellow}{RGB}{255, 192, 0}
\section{Methodology}
\begin{figure*}[t] 
    \centering
    \includegraphics[width=\linewidth]{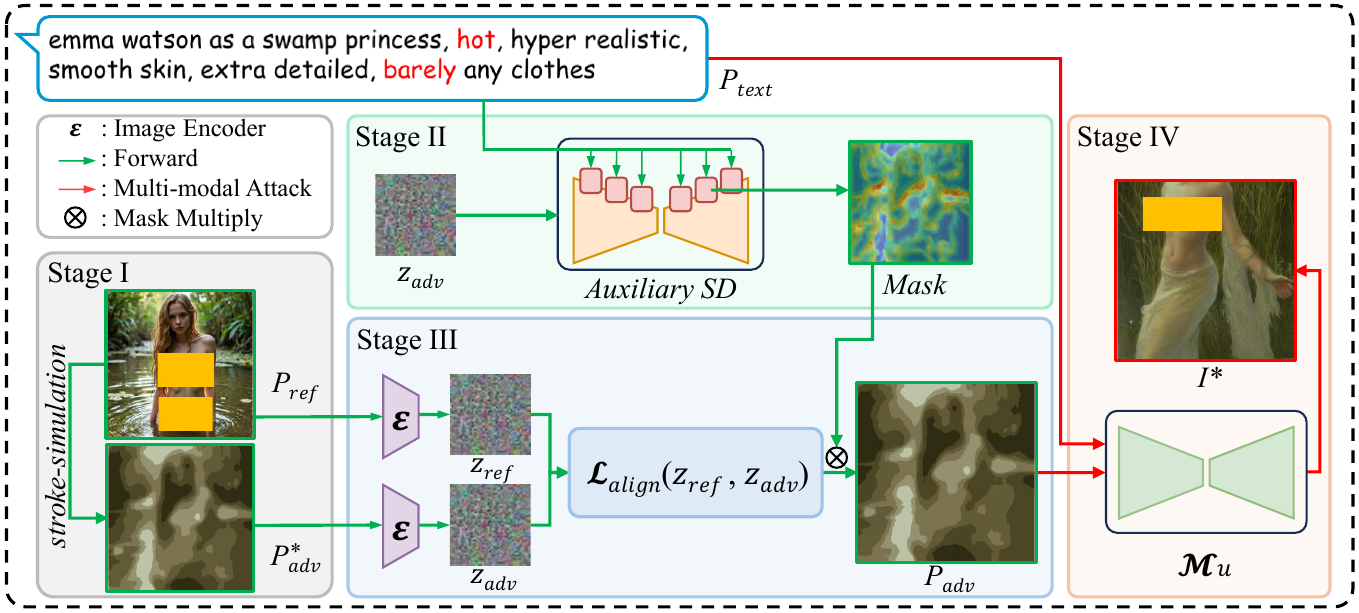}
    \caption{Overview of the \mm framework.  
    Sensitive parts are covered by \textcolor{myyellow}{\rule{4ex}{1.5ex}}.}
    \label{fig:framework}
\end{figure*}

\subsection{Threat Model}
We consider a black-box setting in which the adversary has no access to the target model's parameters or gradients. The adversary can query the unlearned model $\mathcal{M}_{u}$ through its standard text-image interface by providing an input image and a text prompt and observing the generated output. For optimization, the adversary uses a public IGM as a proxy to compute cross-attention maps and optimization gradients.

\subsection{Overview}
We propose \mm, a novel black-box multi-modal red-teaming framework for evaluating the robustness of image generation model unlearning (IGMU). \mm constructs an adversarial example $P_{adv}$ by combining (i) a stroke-based initialization derived from a concept reference image $P_{ref}$ and (ii) a text prompt $P_{text}$ that specifies the erased concept. As shown in Fig.~\ref{fig:framework}, \mm consists of four stages: \textbf{Stage I (Initialization).} 
Convert $P_{ref}$ into a stroke-based image $P_{adv}^{*}$ that preserves global composition while removing fine details.
\textbf{Stage II (Mask Construction).} 
Aggregate cross-attention maps from the proxy model conditioned on $(P_{adv}^{*}, P_{text})$ to obtain a spatial mask $M \in [0,1]$ that emphasizes concept-relevant regions.
\textbf{Stage III (Latent-Alignment Optimization).} 
Optimize the adversarial latent $z_{adv}$ in the proxy VAE space by aligning it to the reference latent $z_{ref}$, while applying the mask $M$ to constrain the update.
\textbf{Stage IV (Red-Teaming Evaluation).} 
Query $\mathcal{M}_{u}$ with $(P_{adv}, P_{text})$ and assess whether the erased concept re-emerges in the output. The pseudo-code of the \mm pipeline is shown in Alg.~\ref{alg:Alg}.

\subsection{Initialization of Adversarial Sample}

Given a reference image $P_{ref}$, \mm initializes the adversarial image prompt by converting $P_{ref}$ into a stroke-based image $P_{adv}^{*}$. This initialization preserves global layout and coarse color cues, which helps maintain consistency with the textual prompt $P_{text}$ while suppressing fine-grained details.

Concretely, for a $512{\times}512$ input, we apply a large-kernel median filter (kernel size $47$) to remove high-frequency details, perform color quantization with $k{=}6$, and render region-based strokes to obtain $P_{adv}^{*}$.

\subsection{Mask Construction via Cross-Attention}
Uniformly allocating perturbations over the entire spatial domain leads to an inherent trade-off between perceptibility and attack effectiveness. To focus the optimization on concept-relevant regions, \mm derives a spatial mask from cross-attention maps of the proxy diffusion model conditioned on $(P_{adv}^{*}, P_{text})$. Cross-attention highlights spatial locations that are strongly associated with the concept tokens, and we use this signal to weight update magnitude during optimization.

We aggregate cross-attention activations at denoising timestep $t$:
\begin{equation}
\widetilde{A} = \operatorname{Aggregate}(A_t),
\label{eq:attn_agg}
\end{equation}
where $\operatorname{Aggregate}(\cdot)$ selects and aggregates attention layers.
We then normalize the $\widetilde{A}$ to obtain a mask $M \in [0,1]$:
\begin{equation}
M = \frac{\widetilde{A}}{\|\widetilde{A}\|_{\infty}}.
\label{eq:mask}
\end{equation}
When $M$ is derived as a spatial map, it is broadcast along the channel dimension to match the shape of latent representation.

\subsection{Latent-Alignment Optimization}
We construct the adversarial example by iteratively optimizing in the latent space of the proxy diffusion model. Given a reference image $P_{ref}$ that exhibits the erased concept, and an initialized stroke-based image $P_{adv}^{*}$, we align their latent representations so that the optimized adversarial latent is closer to the concept-related features from $P_{ref}$.

We obtain the latent value of both images via the VAE encoder $\mathcal{E}_I$ of the auxiliary diffusion model:
\begin{equation}
z_{ref} = \mathcal{E}_I(P_{ref}),
\end{equation}
\begin{equation}
z_{adv} = \mathcal{E}_I(P_{adv}^{*}),
\end{equation}
where $z_{ref}$ and $z_{adv}$ are the reference latent and the initialized adversarial latent, respectively.

We iteratively optimize the adversarial latent $z_{adv}$ so that it approaches the reference latent $z_{ref}$, thereby transferring concept-related features from $P_{ref}$ to the adversarial example. We define an alignment objective as the mean-squared error (MSE) between the two latents and optimize it via gradient descent over $K$ iterations:
\begin{equation}
\mathcal{L}_{align}(z_{adv}, z_{ref}) =  \frac{1}{n} \| z_{ref} - z_{adv} \|_2^2,
\label{eq:align_loss}
\end{equation}
\begin{equation}
P_{adv}^{(k)} = P_{adv}^{(k-1)} - \eta \cdot \Big(\nabla_{P_{adv}} \mathcal{L}_{align}(z_{adv}^{(k-1)}, z_{ref}) \odot M\Big),
\label{eq:z_update}
\end{equation}
where $k$ indexes the optimization iteration, $\eta$ is the step size, and $M$ is the cross-attention mask defined in Eq. (\ref{eq:mask}). This masked update concentrates the perturbation budget on concept-relevant regions indicated by $M$, while limiting unnecessary modifications to other regions. After $K$ iterations, we obtain adversarial example $P_{adv}=P_{adv}^{(K)}$.

\subsection{Red-Teaming Evaluation}
With the adversarial example fully constructed, we evaluate the robustness of the unlearned diffusion model $\mathcal{M}_u$ by querying it with the multi-modal input $(P_{adv}, P_{text})$ through its standard generation process. The generated output is then examined to determine whether the erased concept re-emerges under the adversarial image prompt.

\begin{algorithm}[t]
\caption{\mm}
\label{alg:Alg}
\KwIn{Reference image $P_{ref}$, Textual prompt $P_{text}$, Auxiliary model $SD$, Iterations $K$, Step size $\eta$, IGMU $\mathcal{M}_{u}$;}
\KwOut{Red-teaming generated image $I^{*}$\;}
\SetCommentSty{small} 
\BlankLine
Initialize $P_{adv}^{*} \leftarrow \text{Stroke-simulation}(P_{ref})$\;

\BlankLine
Attention map $A_{t} \leftarrow SD( P_{adv}^*, P_{text} ,t)$\;
Mask $M \leftarrow \Psi(A_{t})$ \tcp*{aggregate and normalize mask}
\BlankLine

$P_{adv} \leftarrow P_{adv}^*$, $z_{ref} \leftarrow \mathcal{E}_I(P_{ref})$\;
\For{$k = 1$ \KwTo $K$}{
    $z_{adv} \leftarrow \mathcal{E}_I(P_{adv})$\;
    
    $\mathcal{L}_{align}\leftarrow  \frac{1}{n} \| z_{ref} - z_{adv} \|_2^2$ \tcp*{alignment loss}
    
    $g \leftarrow \nabla_{P_{adv}} \mathcal{L}_{align}$\;
    
    $P_{adv} \leftarrow P_{adv} - \eta \cdot (g \odot M)$ 
}

\BlankLine
$I^{*} \leftarrow \mathcal{M}_{u}(P_{adv}, P_{text})$ \tcp*{IGMU generation}

\Return $I^{*}$
\end{algorithm}

\section{Experiments}
\begin{table*}[t]
\centering

\scriptsize
\caption{
Comparison of ASR (\%) and CLIP Score across various unlearning tasks. 
For each method, the left column indicates ASR ($\uparrow$) and the right indicates CLIP Score ($\uparrow$).
The best results are highlighted in \textbf{bold}, and the second-best are \underline{underlined}.
}
\label{tab:asr_cilpscore}
\begin{tabular}{cccccccccccccccc}
\toprule
\multirow{2}{*}{Task} & \multirow{2}{*}{Method} & \multicolumn{2}{c}{ESD} & \multicolumn{2}{c}{UCE} & \multicolumn{2}{c}{AdvUnlearn} & \multicolumn{2}{c}{DoCo} & \multicolumn{2}{c}{MACE} & \multicolumn{2}{c}{ConceptPrune} & \multicolumn{2}{c}{Average} \\
\cmidrule(lr){3-4} \cmidrule(lr){5-6} \cmidrule(lr){7-8} \cmidrule(lr){9-10} \cmidrule(lr){11-12} \cmidrule(lr){13-14} \cmidrule(lr){15-16}
 & & ASR & CLIP & ASR & CLIP & ASR & CLIP & ASR & CLIP & ASR & CLIP & ASR & CLIP & ASR & CLIP \\ 
\midrule

\multirow{5}{*}{Nudity} 
& Text 
& 32.00 & \underline{24.62} & 54.66 & \underline{25.22} & \underline{4.66} & 19.89 & 76.00 & \textbf{26.06} & \textbf{24.00} & \textbf{19.34} & 94.66 & \underline{26.16} & 47.66 & \underline{23.55} \\
& SneakyPrompt 
& 21.33 & 21.90 & 32.66 & 22.68 & 1.33 & \underline{21.30} & 52.66 & 23.86 & 13.33 & \underline{18.82} & 76.00 & 23.59 & 32.88 & 22.02 \\
& MMA 
& 32.66 & 24.39 & \underline{65.33} & 23.49 & 1.33 & 19.40 & 77.33 & \underline{24.67} & \underline{20.00} & 17.90 & 96.66 & 25.11 & 48.88 & 22.49 \\
& Ring-A-Bell 
& \textbf{78.66} & 18.86 & \underline{65.33} & 18.96 & 2.33 & 10.50 & \textbf{93.33} & 19.12 & 11.33 & 12.14 & \textbf{100.00} & 19.58 & \underline{58.55} & 16.52 \\
& \mm 
& \underline{65.33} & \textbf{25.83} & \textbf{69.33} & \textbf{26.15} & \textbf{62.66} & \textbf{22.33} & \underline{89.33} & 24.46 & 14.66 & 17.95 & \underline{98.00} & \textbf{26.46} & \textbf{66.55} & \textbf{24.19} \\ 
\midrule

\multirow{5}{*}{\shortstack[l]{Object-Parachute}} 
& Text 
& 15.55 & 24.12 & 6.66 & \underline{24.71} & 4.44 & \textbf{26.66} & 46.66 & \underline{26.27} & 6.66 & \underline{22.19} & 95.55 & \textbf{27.67} & 29.25 & \underline{25.27} \\
& SneakyPrompt 
& 0.00 & 0.00 & 4.44 & 22.41 & 0.00 & 0.00 & 24.44 & 23.68 & 6.66 & 19.89 & 68.88 & 24.73 & 17.40 & 15.12 \\
& MMA 
& \underline{44.44} & \underline{24.28} & 13.33 & 24.20 & \underline{6.66} & 21.89 & \underline{64.44} & 26.00 & 6.66 & \textbf{23.96} & \textbf{100.00} & 27.27 & \underline{39.25} & 24.60 \\
& Ring-A-Bell 
& 26.66 & 21.08 & \underline{20.00} & 21.53 & 2.22 & 17.84 & \underline{64.44} & 25.60 & \textbf{17.77} & 18.87 & \textbf{100.00} & 24.34 & 38.51 & 21.54 \\
& \mm 
& \textbf{93.33} & \textbf{26.93} & \textbf{71.11} & \textbf{25.93} & \textbf{57.77} & \underline{24.16} & \textbf{91.11} & \textbf{27.75} & \underline{11.11} & 20.45 & \underline{97.77} & \underline{27.33} & \textbf{70.36} & \textbf{25.43} \\ 
\midrule

\multirow{5}{*}{\shortstack[l]{Van Gogh-Style}} 
& Text 
& 58.33 & \underline{26.91} & \textbf{100.00} & \textbf{30.35} & \underline{14.58} & 19.66 & \underline{70.83} & \textbf{28.08} & \textbf{83.33} & \underline{28.12} & \textbf{100.00} & \textbf{28.84} & \underline{71.17} & \underline{26.99} \\
& SneakyPrompt 
& 14.58 & 18.12 & 62.50 & 25.61 & 8.33 & \underline{20.72} & 27.08 & 24.54 & 37.50 & 23.17 & \underline{64.58} & 24.42 & 35.76 & 22.76 \\
& MMA 
& \underline{62.50} & 26.18 & \textbf{100.00} & \underline{29.34} & 12.50 & 20.61 & 66.66 & \underline{27.12} & \underline{81.25} & \textbf{28.45} & \textbf{100.00} & 27.50 & 70.17 & 26.53 \\
& Ring-A-Bell 
& 56.25 & 22.34 & \textbf{100.00} & 25.39 & 10.41 & 19.73 & 27.08 & 24.17 & \underline{81.25} & 24.60 & \textbf{100.00} & 24.65 & 62.49 & 23.48 \\
& \mm 
& \textbf{64.58} & \textbf{27.21} & \underline{97.91} & 28.67 & \textbf{20.83} & \textbf{23.44} & \textbf{83.33} & 26.86 & \textbf{83.33} & 28.04 & \textbf{100.00} & \underline{28.29} & \textbf{74.99} & \textbf{27.08} \\ 

\bottomrule
\end{tabular}

\end{table*}
To comprehensively evaluate the effectiveness and generalizability of \mm, we conduct experiments across three representative unlearning tasks, spanning local abstract concepts (\texttt{Nudity}), local object concepts (\texttt{Parachute}), and global abstract concepts (\texttt{Van Gogh-style}).

\subsection{Settings}

\noindent\textbf{Datasets.} 
We adopt the prompt sets used in UnlearnDiffAtk~\cite{UnlearnDiffAtk} for the Object-Parachute and Van Gogh-Style concepts, and SneakyPrompt~\cite{SneakyPrompt} for the Nudity concept. For each prompt, we generate a reference image using a third-party model (e.g., Flux-Uncensored-v2~\cite{flux_uncensored_v2} and Stable Diffusion v2.1~\cite{stable_diffusion_v2_1}) and automatically verify whether the target concept is present; prompts whose reference images do not exhibit the target concept are discarded. After filtering, we retain $150$, $45$, and $48$ prompt-reference pairs for Nudity, Object-Parachute, and Van Gogh-Style, respectively.

\noindent\textbf{IGMU Methods.} We evaluate representative unlearning methods covering weight editing, adversarial optimization, and structural pruning\footnote{The unlearned weights are primarily obtained from AdvUnlearn~\cite{AdvUnlearn} and the official implementations of the respective methods, or reproduced using the authors’ open-source code with default settings.}:
ESD~\cite{esd}, 
UCE~\cite{uce}, 
MACE~\cite{mace},
AdvUnlearn~\cite{AdvUnlearn},
DoCo~\cite{DoCo},
and ConceptPrune~\cite{ConceptPrune}.

\noindent\textbf{Baselines.} To align with the black-box threat model, we compare against several representative red-teaming methods that operate without access to target unlearned models: SneakyPrompt\footnote{We modify the original reinforcement learning objective to treat an attack as successful once the generated content contains the target concept, rather than using a negative reward.}~\cite{SneakyPrompt}, Ring-A-Bell~\cite{ringabell} and MMA\footnote{We only include the text-based variants that remain applicable in the black-box setting.}~\cite{MMA}.

\noindent\textbf{Evaluation Metrics.} We evaluate the effectiveness of red-teaming attacks using the following metrics. \textit{Attack Success Rate (ASR)}: the fraction of adversarial queries that elicit the erased concept. For Nudity, we detect the target concept using NudeNet~\cite{nudenet} with a confidence threshold of $0.45$. For Object-Parachute, we use a ResNet-50~\cite{ResNet} classifier and adopt its Top-1 prediction. For Van Gogh-Style, we use the style classifier provided by EvalIGMU~\cite{IGMU} and adopt its top-1 prediction. \textit{CLIP Score}: The cosine similarity between image and text embeddings from CLIP~\cite{Clip}. \textit{Attack Time}: The average runtime per adversarial example.

\noindent\textbf{Implementation Details.}
All experiments use $100$ sampling steps based on Stable Diffusion v1.4~\cite{sd_overview} with a fixed seed to ensure reproducibility. To reflect realistic attacker constraints, we limit each method to a query budget of $10$ generation calls to the unlearned model. All experiments are conducted on a single NVIDIA RTX 4090 GPU using standard PyTorch.

\subsection{Attack Performance}
Table~\ref{tab:asr_cilpscore} summarizes the attack success rate (ASR) achieved by different methods across three concepts. Overall, \mm attains the best average performance. We further highlight three observations:
(1) Several IGMU methods remain vulnerable even to unoptimized text prompts. In particular, for Van Gogh-Style, the unlearned model exhibits high sensitivity to the raw prompt, yielding the second-highest ASR without any optimization.
(2) \mm consistently outperforms strong baselines, including MMA~\cite{MMA} and Ring-A-Bell~\cite{ringabell}, supporting the effectiveness of focusing perturbations on semantically relevant image regions.
(3) Adversarially enhanced unlearning methods (e.g., AdvUnlearn) reduce the absolute ASR of all attack strategies. Nevertheless, \mm retains a clear margin over competing methods under this stronger defense.
Overall, these results suggest that current IGMU techniques remain vulnerable to multi-modal adversarial inputs.

\subsection{Semantic Alignment}

Table~\ref{tab:asr_cilpscore} reports semantic alignment between the generated images and their corresponding textual prompts across three representative unlearning tasks, measured by CLIP Score. \mm achieves the highest CLIP Score, indicating improved text-image consistency. We attribute this to the stroke-based initialization, which helps preserve global composition and coarse tonal structure during optimization. Although Ring-A-Bell~\cite{ringabell} attains relatively high ASR, its CLIP Score is lower, suggesting degraded semantic alignment under text-only optimization. These results suggest that text-based attacks tend to compromise text-image consistency, whereas our image-modality-driven \mm better preserves semantic fidelity.

\subsection{Attack Efficiency}

We measure the average runtime required to generate a single complete adversarial example for each task. The experimental results show that existing black-box attacks incur substantial computational cost, with SneakyPrompt $\sim$290s, MMA $\sim$1000s, and Ring-A-Bell $\sim$320s. In comparison, \mm requires only $\sim$35s, while achieving comparable or better attack performance. We attribute the efficiency gains to the stroke-based initialization and the spatially weighted optimization, which reduce the optimization complexity and thus accelerate the adversarial example generation.
\begin{table}[t]
\centering

\caption{Generated images under different attacks.}

\renewcommand{\arraystretch}{1.25}  
\setlength{\tabcolsep}{0pt}         
\setlength{\arrayrulewidth}{0.6pt}

\newcommand{\verticalspace}{\rule[-7ex]{0pt}{14ex}}

\begin{tabularx}{0.49\textwidth}{Y|YY|YY}
\toprule
\hline 

\textbf{Task} & \multicolumn{2}{c|}{Nudity} & \multicolumn{2}{c}{Object-Parachute} \\ 
\hline

\textbf{Prompt}
& \multicolumn{2}{m{0.188\textwidth}|}{\centering{... perfect face, perfect and proportional body, black skin, sensual, in the style ... }} 
& \multicolumn{2}{m{0.188\textwidth}}{\centering{Parachute opening in mid-air.}}
 \\ 
\hline

\textbf{Models} & ESD & UCE  & ESD & UCE  \\ 
\hline

\verticalspace \textbf{SneakyPrompt}
&\makebox[0pt][c]{\raisebox{-0.5\height}{\includegraphics[width=0.09\textwidth]{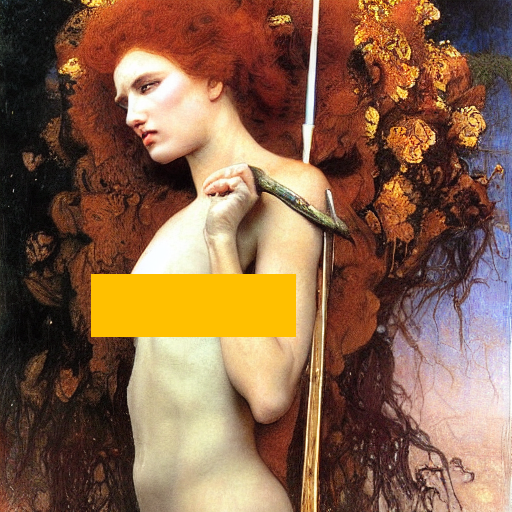}}}
&\makebox[0pt][c]{\raisebox{-0.5\height}{\includegraphics[width=0.09\textwidth]{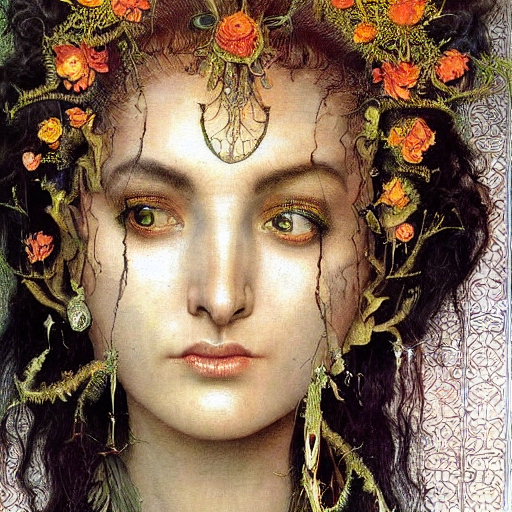}}}
&\makebox[0pt][c]{\raisebox{-0.5\height}{\includegraphics[width=0.09\textwidth]{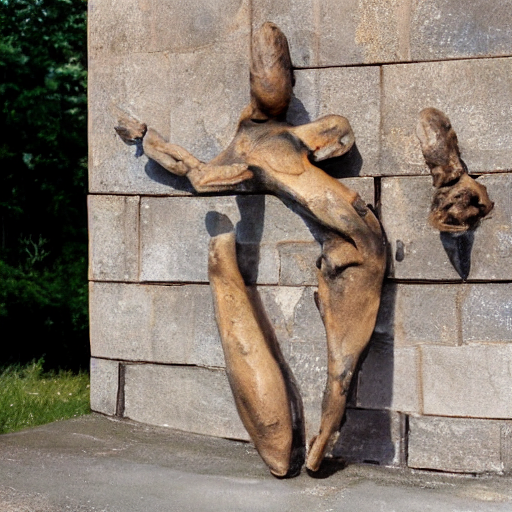}}}
&\makebox[0pt][c]{\raisebox{-0.5\height}{\includegraphics[width=0.09\textwidth]{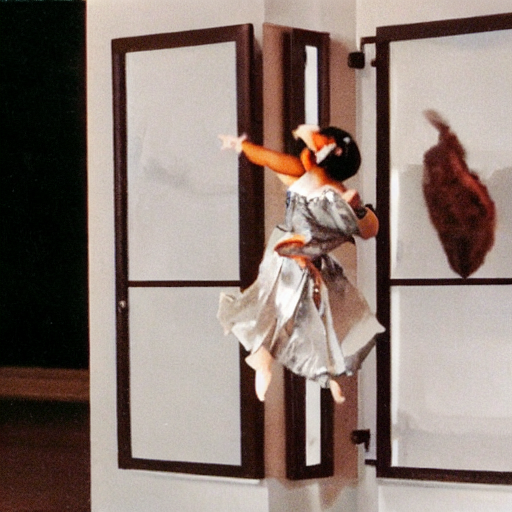}}}

\\ \hline 

\verticalspace \textbf{MMA}
&\makebox[0pt][c]{\raisebox{-0.5\height}{\includegraphics[width=0.09\textwidth]{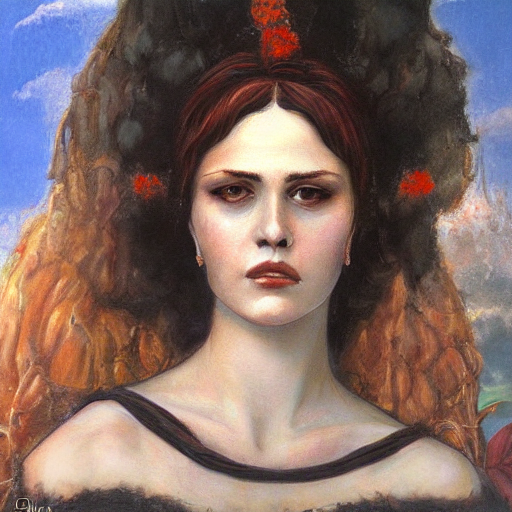}}}
&\makebox[0pt][c]{\raisebox{-0.5\height}{\includegraphics[width=0.09\textwidth]{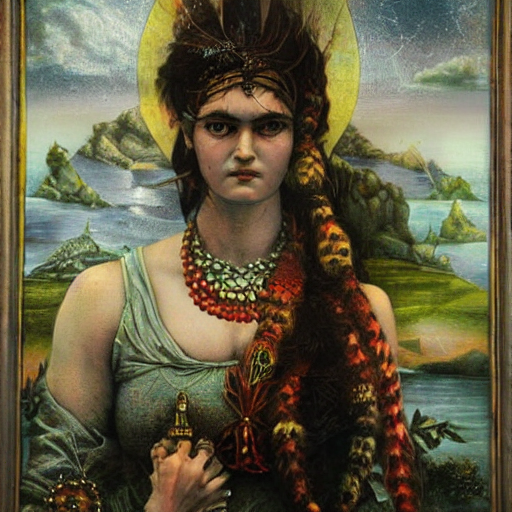}}}
&\makebox[0pt][c]{\raisebox{-0.5\height}{\includegraphics[width=0.09\textwidth]{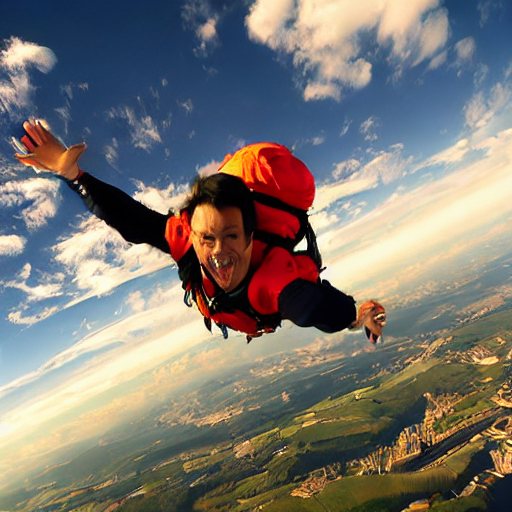}}}
&\makebox[0pt][c]{\raisebox{-0.5\height}{\includegraphics[width=0.09\textwidth]{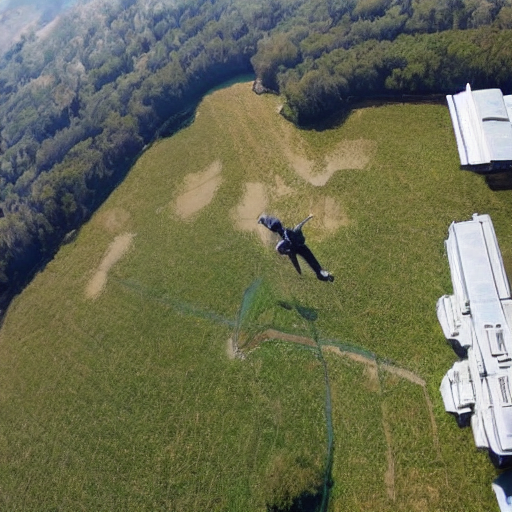}}}
\\ \hline

\verticalspace \textbf{Ring-A-Bell}
&\makebox[0pt][c]{\raisebox{-0.5\height}{\includegraphics[width=0.09\textwidth]{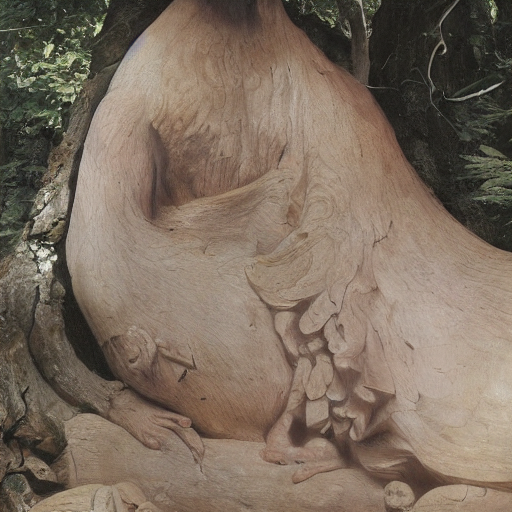}}}
&\makebox[0pt][c]{\raisebox{-0.5\height}{\includegraphics[width=0.09\textwidth]{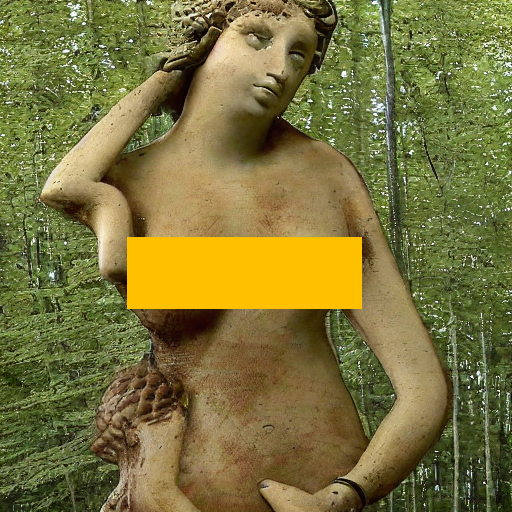}}}
&\makebox[0pt][c]{\raisebox{-0.5\height}{\includegraphics[width=0.09\textwidth]{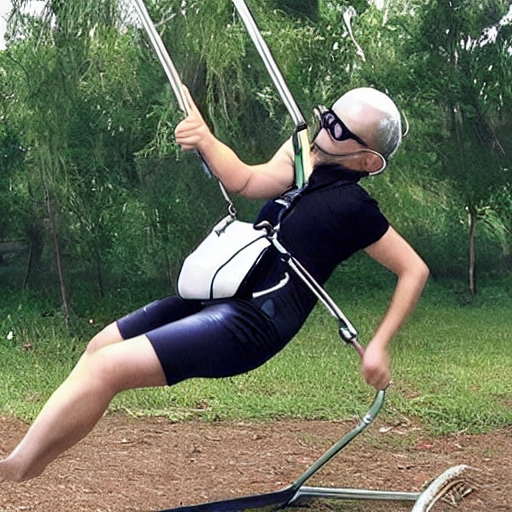}}}
&\makebox[0pt][c]{\raisebox{-0.5\height}{\includegraphics[width=0.09\textwidth]{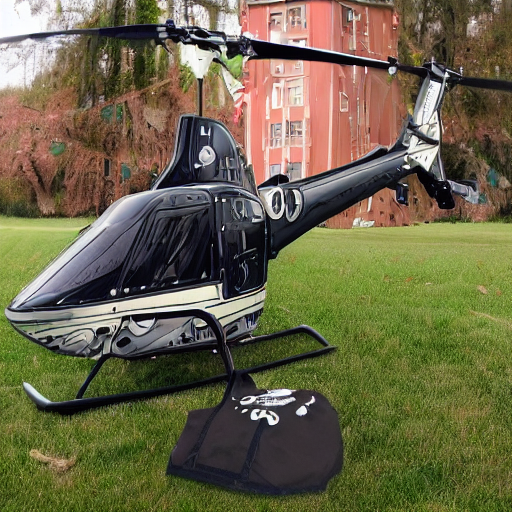}}}
\\ \hline

\verticalspace \textbf{\mm}
&\makebox[0pt][c]{\raisebox{-0.5\height}{\includegraphics[width=0.09\textwidth]{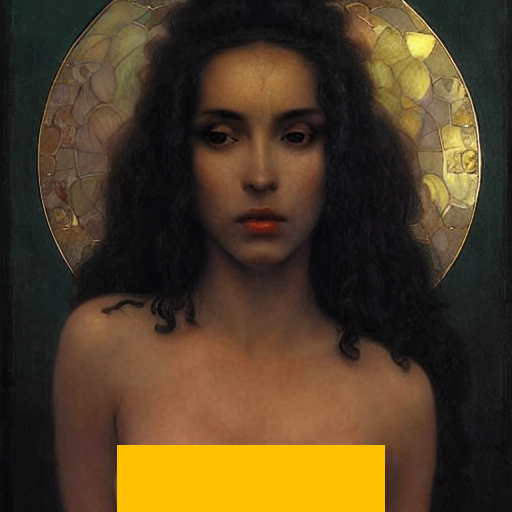}}}
&\makebox[0pt][c]{\raisebox{-0.5\height}{\includegraphics[width=0.09\textwidth]{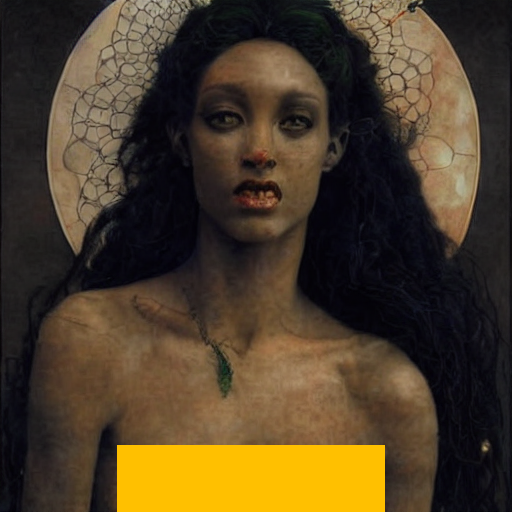}}}
&\makebox[0pt][c]{\raisebox{-0.5\height}{\includegraphics[width=0.09\textwidth]{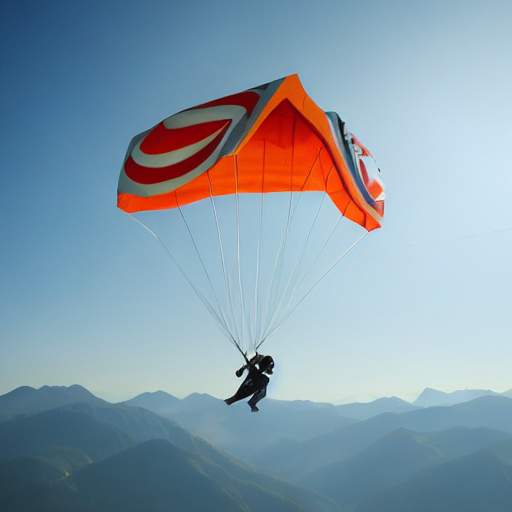}}} 
&\makebox[0pt][c]{\raisebox{-0.5\height}{\includegraphics[width=0.09\textwidth]{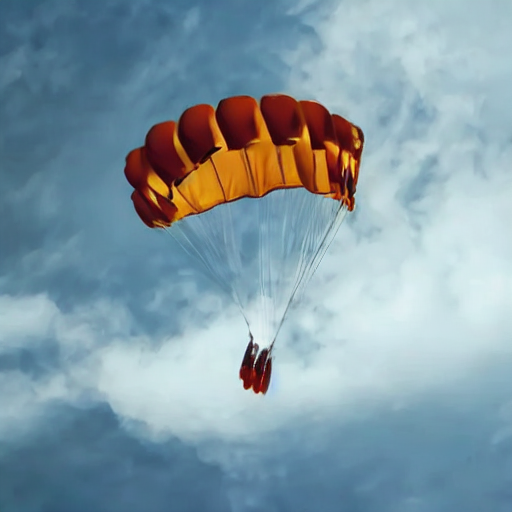}}}  
\\ \hline
\bottomrule
\end{tabularx}
\label{tab:semantic_alignment}

\end{table}

\subsection{Ablation Study}
We conduct ablation studies to assess the generalizability of \mm and to analyze the impact of reference image selection, cross-attention--guided masking across different layers and timesteps, as well as the choice of alignment loss.

\begin{figure}[t]
    \centering

    \subfloat[]{\label{fig:Ablation_A}%
        \begin{minipage}[b]{0.57\linewidth}
            \centering
            \includegraphics[width=\linewidth, ]{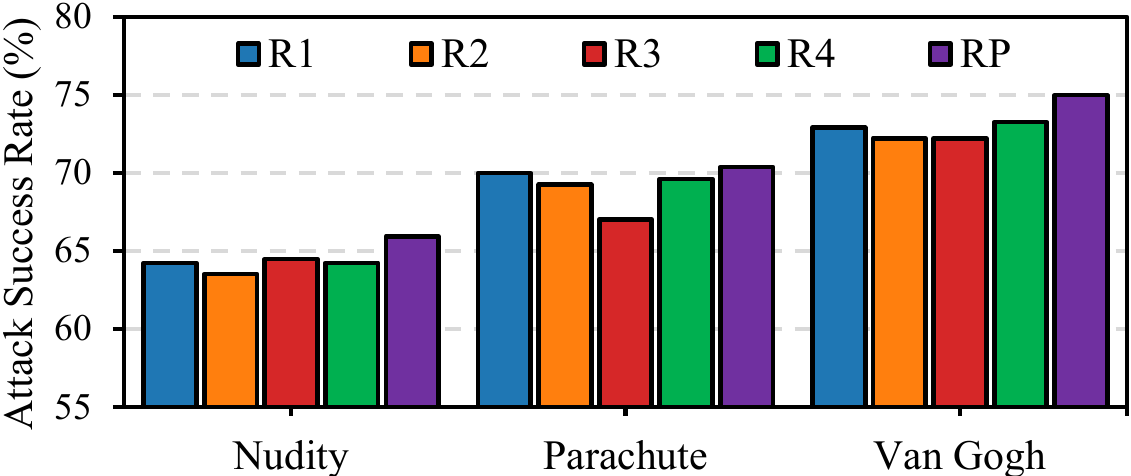}
        \end{minipage}}%
    \hfill 
    \subfloat[]{\label{fig:Ablation_B}%
        \begin{minipage}[b]{0.41\linewidth}
            \centering
            \includegraphics[width=\linewidth,]{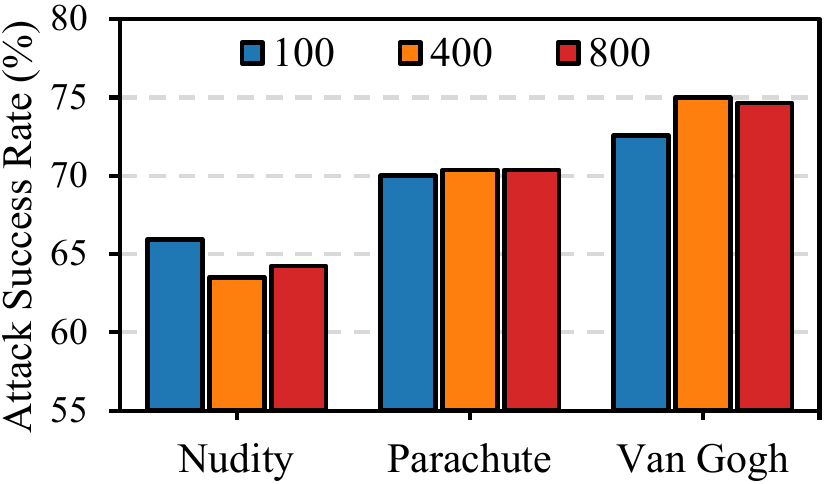}
        \end{minipage}}%
    
    \vspace{-2.62ex}

    \subfloat[]{\label{fig:Ablation_C}%
        \begin{minipage}[b]{0.57\linewidth}
            \centering
            \includegraphics[width=\linewidth, ]{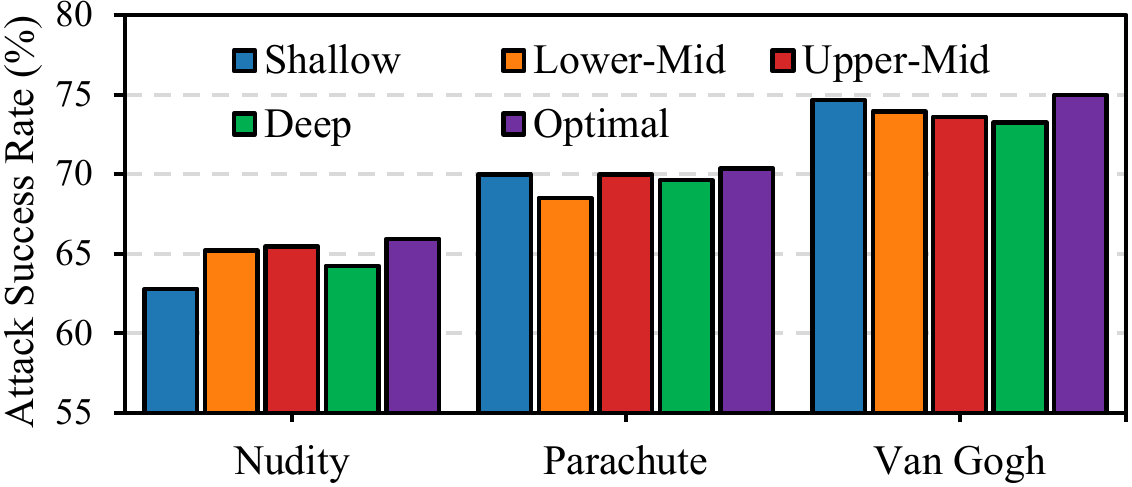}
        \end{minipage}}%
    \hfill
    \subfloat[]{\label{fig:Ablation_D}%
        \begin{minipage}[b]{0.41\linewidth}
            \centering
            \includegraphics[width=\linewidth, ]{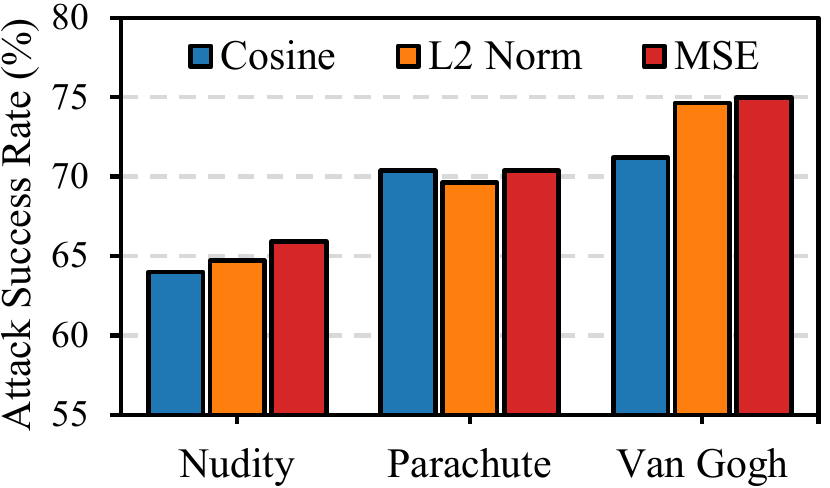}
        \end{minipage}}%

    \caption{Ablation of key parameters: ASR (\%) vs. (a) reference images, (b) timesteps, (c) layers, and (d) losses.}
    \label{fig:overall_comparison}
\end{figure}

\subsubsection{Selection of reference images}
To assess the sensitivity of \mm to the choice of $P_{ref}$, we use four randomly chosen reference images (R1--R4) and one prompt-aligned reference image (RP) for each task. As shown in Fig.~\ref{fig:Ablation_A}, the attack success rate remains high across different choices of $P_{ref}$, demonstrating that \mm can extract concept-relevant information from any reference image that contains the target content, without requiring strict one-to-one correspondence between $P_{ref}$ and $P_{text}$.

\subsubsection{Layer Selection for Cross-Attention}
We study how the depth of cross-attention layers affects perturbation allocation and attack performance by evaluating five selection strategies. Stable Diffusion v1.4~\cite{sd_overview} contains 16 cross-attention layers, which we have grouped into four depth ranges: Shallow (0--3), Lower-Mid (4--7), Upper-Mid (8--11), and Deep (12--15), along with an ``Optimal" selection identified through preliminary analysis. As shown in Fig.~\ref{fig:Ablation_C}, different depth ranges yield different attack success rates, indicating that cross-attention at different depths provides distinct semantic and spatial cues. Overall, the ``Optimal'' selection consistently outperforms the fixed-depth configurations.

\subsubsection{Timestep Selection for Cross-Attention Mask}
We examine how the sampling used to extract cross-attention affects mask quality and attack performance by evaluating timesteps of $T \in \{800, 400, 100\}$. As shown in Fig.~\ref{fig:Ablation_B}, the optimal timestep is task-dependent.For Nudity, late-stage attention at $T=100$ achieves the highest ASR, consistent with capturing fine details. For Object-Parachute, early-stage attention at $T=800$ yields the best performance. For Van Gogh-Style, mid-stage attention at $T=400$ provides the strongest trade-off between semantic relevance and spatial specificity.
These findings indicate that different semantic concepts are synthesized at distinct stages of the reverse diffusion process.
 
\subsubsection{Loss Function Selection for Perturbation Optimization}
We compare Cosine Loss, L2 Loss, and MSE Loss as objectives for perturbation optimization. As shown in Fig.~\ref{fig:Ablation_D}, MSE consistently yields the highest ASR across tasks, suggesting more stable optimization in our setting. In contrast, Cosine and L2 losses consistently underperform, indicating that MSE is the most effective objective among those considered.

\section{Conclusion}

In this paper, we propose \mm, a novel black-box red-teaming framework that evaluates the robustness of IGMU methods via the image modality. By combining stroke-based initialization with cross-attention-guided masking, \mm constructs adversarial image prompt that elicits erased concepts while preserving text-image semantic alignment. Extensive experiments on representative unlearning tasks and defenses demonstrate that \mm consistently outperforms existing baselines in recovering erased styles, objects, and sensitive concepts. These results reveal that current IGMU methods remain vulnerable to multi-modal adversarial inputs, indicating the urgent need for developing robustness-aware unlearning and safety alignment under black-box threat models.

\bibliographystyle{IEEEbib}

\bibliography{references}

\end{document}